%% file: main.tex
\documentclass[11pt]{article}
\usepackage[final]{acl}
\input{config}

\title{Code over Words: \\ Overcoming Semantic Inertia via Code-Grounded Reasoning}

\author{
    \textbf{Manjie Xu\textsuperscript{1,2,5,6*}}\quad
    \textbf{Isabella Yin\textsuperscript{3*}}\quad
    \textbf{Xinyi Tu\textsuperscript{4}}\quad
    \textbf{Chi Zhang\textsuperscript{1,5\,\,\textrm{\Letter}}}\quad
    \textbf{Yixin Zhu\textsuperscript{2,1,5,6,7\,\,\textrm{\Letter}}}
    \\
    \small\textsuperscript{1} Institute for Artificial Intelligence, Peking University\quad
    \small\textsuperscript{2} School of Psychological and Cognitive Sciences, Peking University\\
    \small\textsuperscript{3} Tsinghua International School\quad    \small\textsuperscript{4} University of California, Berkeley\quad
    \small\textsuperscript{5} State Key Lab of General AI, Peking University\\
    \small\textsuperscript{6} Beijing Key Laboratory of Behavior and Mental Health, Peking University\\
    \small\textsuperscript{7} Embodied Intelligence Lab, PKU-Wuhan Institute for Artificial Intelligence\\
    \small * equal contribution\quad \textrm{\Letter} correspondence authors:\,\texttt{chizhang.cz@pku.edu.cn}\,\,\texttt{yixin.zhu@pku.edu.cn}
}

\begin{document}
\maketitle

\begin{abstract}
\acp{llm} struggle with \textit{Semantic Inertia}: the inability to inhibit pre-trained priors (\eg, ``Lava is Dangerous'') when dynamic, in-context rules contradict them. 
We probe this phenomenon using \game, where physical laws are mutable text rules, enabling precise evaluation of models’ ability to override learned priors when rules change. We quantatively observe that larger models can exhibit \textit{inverse scaling}: they perform worse than smaller models when natural language reasoning requires suppressing pre-trained associations (\eg, accepting ``Lava is Safe'').
Our analysis attributes this to natural language encoding, which entangles descriptive semantics and logical rules, leading to persistent hallucinations of familiar physics despite explicit contradictory rules.
Here we show that representing dynamics as executable code, rather than descriptive text, reverses this trend and enables effective prior inhibition.
We introduce \acf{lcv}, which fine-tunes models on counterfactual pairs and identifies states with contradictory rules, thereby forcing attention to logical constraints rather than visual semantics. This training-time approach outperforms expensive inference-time search methods in both efficiency and accuracy.
Our results demonstrate that representation fundamentally determines whether scaling improves or impairs contextual reasoning. This challenges the assumption that larger models are universally better, with implications for domains that require dynamic overriding of learned priors.
\end{abstract}

\section{Introduction}

\acfp{llm} have demonstrated remarkable reasoning capabilities by leveraging vast commonsense knowledge embedded in their pre-training data \citep{brown2020language, bubeck2023sparks}. However, this success relies fundamentally on distributional semantics---the statistical correlation between words and concepts learned from text corpora \citep{harris1954distributional,mikolov2013distributed}. As \citet{bender2020climbing} argue, models trained purely on linguistic form struggle when meaning must be decoupled from statistical priors. While efficient for static domains, robust intelligence demands the ability to dynamically reassign meaning based on context \citep{lake2017building, zhu2020dark}. When told ``Lava is Safe,'' an agent must override the deeply ingrained association with danger and act on the explicit logical rule instead. We term this fundamental limitation \textit{Semantic Inertia}---the tendency for parametric memory to override in-context reasoning.

\begin{figure}[t!]
    \centering
    \includegraphics[width=\linewidth]{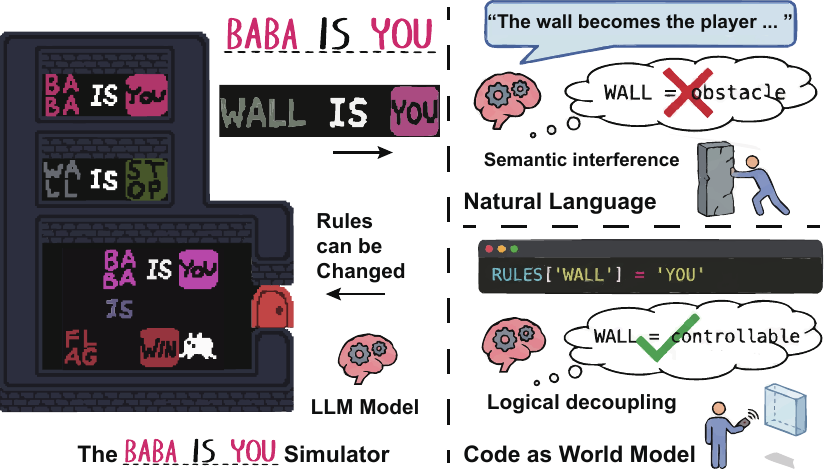}
    \caption{\textbf{Rule mutability and reasoning paradigms in \game.} The \game environment externalizes game logic as manipulable text blocks, allowing the rules governing object affordances and agent identity to be dynamically rewritten (\eg, ``Baba is You'' $\rightarrow$ ``Wall is You''). We study two reasoning paradigms in this setting: natural language reasoning, which operates over descriptive semantics and is prone to semantic inertia, and code-grounded reasoning, which treats rules as executable constraints and enables explicit state tracking under mutable laws.}
    \label{fig:intro}
\end{figure}

We rigorously evaluate this capability using \game\footnote{\url{https://github.com/utilForever/baba-is-auto}}, a logic puzzle environment where physical laws are mutable text blocks. Unlike traditional reinforcement learning domains such as Minecraft \citep{fan2022minedojo} where object affordances remain fixed, \game collapses the boundary between object interactions and physical laws. Rules exist as tangible ``text blocks'' within the game world; manipulating them enables ontological restructuring, transforming language from descriptive labels into causal operators \citep{pearl2009causality} (\cref{fig:intro}). This design creates a stringent test of non-monotonic reasoning \citep{mccarthy1980circumscription}: any inference remains valid only while the current rule configuration persists, and that configuration itself is subject to change.

This environment induces cognitive conflicts analogous to well-studied phenomena in human psychology. Functional fixedness \citep{duncker1945problem} describes the difficulty of reassigning familiar objects to novel functions, while inhibitory control \citep{diamond2013executive} measures the capacity to suppress prepotent responses. \game creates a computational analog of the Stroop effect \citep{stroop1935studies}: perceiving a \texttt{WALL} sprite triggers the ``Obstacle'' concept, yet the rule \texttt{WALL IS YOU} demands the contradictory assignment ``Avatar.'' Successful reasoning requires sustained inhibition of parametric priors in favor of explicit contextual constraints.

Our experiments reveal that state-of-the-art \acp{llm} struggle profoundly with such conflicts. Both standard prompting and reasoning techniques like \ac{cot} \citep{wei2022chain} fail to overcome semantic inertia, causing models to hallucinate solutions based on pre-trained priors---even after explicitly acknowledging contradictory rules in context \citep{mckenzie2023inverse,turpin2023language,stringli2025pitfalls}. This aligns with recent findings that \acp{llm} exhibit systematic biases favoring parametric knowledge over contextual updates \citep{jiang2023mewl,zhaounderstanding,yamin2025llms}. Counterintuitively, we observe that larger models perform worse than smaller ones on high-conflict reasoning tasks, demonstrating inverse scaling where increased capacity amplifies rather than mitigates semantic inertia \citep{mckenzie2023inverse}.

This phenomenon reveals a fundamental misalignment in text-based reasoning architectures \citep{stringli2025pitfalls}. Autoregressive language models predict tokens through surface-level statistical patterns rather than enforcing global consistency across evolving state spaces \citep{yao2023tree}. Without explicit state representations and verifiable transition functions, models resolve ambiguous or counterintuitive rules through plausibility heuristics rather than causal derivation, leading to hallucinated outputs that conform to entrenched priors instead of rigorous logical inference \citep{shinn2023reflexion}.

To address this limitation, we propose \acf{lcv}, which enforces explicit state tracking by grounding reasoning in executable code rather than natural language \citep{liang2023code,gao2023pal,ahmed2025synthesizing}. Building on recent code-as-planner approaches \citep{gao2023pal,tang2024worldcoder,wong2024learning}, \ac{lcv} reframes the \ac{llm}'s role from passive sequence generator to active theorist: instead of directly predicting actions, the model synthesizes Python programs that instantiate the current ``laws of physics'' governing the environment \citep{chen2022program}. Crucially, unlike inference-time methods such as TheoryCoder \citep{ahmed2025synthesizing} that require expensive iterative generate--test--debug cycles, \ac{lcv} performs \textit{amortized theory induction}. Through supervised fine-tuning on counterfactual contrastive pairs---identical states governed by contradictory rules---the model learns to synthesize correct world models in a single forward pass, explicitly disentangling visual appearance from logical affordances and suppressing semantic priors in favor of context-dependent dynamics.

Our contributions are threefold:
(i) \textbf{Inverse Scaling Analysis.} We provide the first quantitative demonstration of inverse scaling in rule-following tasks, showing that without code grounding, larger models exhibit significantly stronger semantic inertia than smaller counterparts \citep{mckenzie2023inverse}.
(ii) \textbf{Amortized \ac{lcv} Framework.} We introduce \ac{lcv} for amortized theory induction through counterfactual contrastive alignment. This training-time intervention enables single-pass synthesis of executable world models, breaking semantic inertia by structurally decoupling logical dynamics from parametric priors.
(iii) \textbf{Efficiency and Robustness.} We demonstrate that \ac{lcv} outperforms strong inference-time search baselines on flexible rule reasoning while greatly reducing inference latency, proving that overcoming semantic inertia requires representational alignment rather than increased computational budget.

\section{Related Work}

\paragraph{\texorpdfstring{\ac{llm}}{} Reasoning and Embodied Planning}
Recent advances have extended \acp{llm} beyond static question answering toward embodied agentic planning \citep{xu2023active,xi2025rise,acharya2025agentic}. Techniques like \ac{cot} \citep{wei2022chain} and Tree of Thoughts \citep{yao2023tree} decompose complex goals into intermediate reasoning steps, while open-world agents such as Voyager \citep{wang2023voyager} and GITM \citep{zhu2023ghost} leverage iterative prompting for skill discovery in Minecraft. However, these environments assume static physics: fundamental object affordances (\eg, ``Wall is Stop'') remain immutable constants throughout interaction. Our work addresses a fundamentally different challenge---dynamic ontology, where an agent must reason under axioms that function as mutable variables rather than fixed priors. While text-based games have been explored \citep{shridhar2020alfworld,yao2022react,li2024chain}, they rarely demand inhibition of strong semantic associations (\eg, accepting ``Lava is Safe''). Recent studies confirm that planning success in static environments does not transfer to domains requiring ontological restructuring \citep{yamin2025llms,van2025baba}.

\paragraph{Semantic Inertia and Inverse Scaling}
A critical limitation of foundation models is their tendency to prioritize parametric knowledge over contextual information---a phenomenon termed semantic inertia or prior bias. \citet{bian2024chatgpt} and \citet{wu2024reasoning} show that \ac{llm} performance degrades substantially on counterfactual reasoning tasks (\eg, ``gravity acts upwards''). More troublingly, \citet{mckenzie2023inverse} identify inverse scaling: larger models, having absorbed more human-centric training data, become \textit{harder} to steer away from commonsense priors than their smaller counterparts. This aligns with theoretical frameworks characterizing hallucination as conflict between bottom-up input and top-down parametric memory \citep{zhang2025siren}. Existing mitigation strategies---self-consistency \citep{wang2022self}, multi-agent debate \citep{du2023improving}---operate within natural language, yet we argue this medium itself introduces ambiguity: soft attention mechanisms inherently struggle to enforce strict inhibition of semantic associations. Following recent work \citep{tangworldcoder,xuheterogeneous,ahmed2025synthesizing}, we ground reasoning in executable code to impose discrete overrides of linguistic intuition.

\paragraph{Code-Based World Modeling}
Code-as-reasoning has emerged as a paradigm for decoupling logic from linguistic ambiguity. Methods like \ac{pal} \citep{gao2023pal} and Program of Thoughts \citep{chen2022program} delegate arithmetic tasks to Python interpreters, while ViperGPT \citep{suris2023vipergpt} synthesizes programs for visual reasoning. Most relevant to our work are frameworks synthesizing explicit world models for planning: WorldCoder \citep{tangworldcoder}, Chain-of-Code \citep{li2024chain}, and TheoryCoder \citep{ahmed2025synthesizing} enable \acp{llm} to construct low-level transition dynamics supporting high-level planning. However, these approaches \citep{liu2023llm+,wong2023word,das2023combining} rely predominantly on inference-time search through iterative generate-test-debug loops. Works like Ada \citep{wong2024learning} automatically construct task-specific planning representations but inherit similar computational costs.

We identify \textbf{two fundamental limitations}. First, inference-time search scales linearly with verification complexity, imposing prohibitive latency for real-time deployment. Second, iterative debugging exhibits confirmation bias under strong priors: models may reject valid counterintuitive rules because they conflict with entrenched semantic heuristics. The proposed \ac{lcv} addresses both issues through \textit{amortized theory induction}. Specifically, rather than searching for theories at test time, we employ counterfactual contrastive alignment during training to internalize the capability for single-pass synthesis. This reduces inference latency by 4 times while improving robustness to semantic interference, demonstrating that overcoming inertia requires representational alignment rather than increased computational budget.

\section{\benchmark: Evaluating Ontological Plasticity}

To rigorously disentangle semantic inhibition failures from general planning deficits, we introduce \benchmark, a benchmark designed to test whether models can suppress entrenched semantic associations and correctly apply dynamic, counterintuitive rules. While prior work on \game \citep{charity2022keke,cloos2024baba} provides foundational testbeds, these environments either lack sufficient challenge or fail to provide paired comparisons necessary for isolating specific failure modes. Recent work confirms that \game remains challenging for state-of-the-art models and continues to serve as a valuable testbed for reasoning under dynamically changing ontologies \citep{van2025baba}.

\benchmark extends these foundations by treating physical laws as latent, mutable variables rather than fixed constants. Each problem instance is a tuple $M_t = \langle S_t, R_t, V \rangle$, where $S_t$ denotes the grid state and $R_t$ represents currently active logic rules derived directly from text block arrangements in $S_t$. Critically, the transition function $T: S \times A \rightarrow S'$ is not static but parameterized by the current rule set: $T(\cdot\ ; R_t)$. Success requires reconstructing underlying ontological concepts: agents must manipulate text blocks to rewrite $R_t$ (\eg, redefining a solid ``Wall'' as passable), thereby transforming language from descriptive label into causal operator.

We procedurally generate levels across three tiers of increasing semantic dissonance, inspired by the \textit{Stroop Test} paradigm (see \cref{sec:supp:env_viz}): (i) \textbf{Semantic Alignment}---rules align with pre-training priors (\eg, \texttt{WALL IS STOP}, \texttt{KEY IS OPEN}), establishing baseline spatial reasoning; (ii) \textbf{Semantic Conflict}---rules violate commonsense physics (\eg, \texttt{LAVA IS SAFE}, \texttt{WALL IS YOU}), creating Stroop-like conflicts requiring inhibition of parametric priors; (iii) \textbf{Dynamic Plasticity}---multi-stage levels where rules shift mid-episode (\eg, constructing \texttt{WALL IS PASS} to escape, then dismantling it to block pursuit), testing dynamic semantic updating without rule persistence.

\section{The Inverse Scaling of Semantic Inertia}

Before introducing our \ac{lcv} pipeline, we establish its empirical motivation through a probing analysis of a critical question: \textit{Does scaling model size automatically resolve semantic inertia?}

While reasoning capabilities generally improve with scale, recent work shows that certain biases can actually worsen---a phenomenon termed \textit{inverse scaling} \citep{mckenzie2023inverse, yamin2025llms, stringli2025pitfalls}. We hypothesize this occurs in \game because larger models develop more entrenched distributional priors about concepts like ``Wall'' or ``Lava.'' In contrast, code-based state representations may decouple logical operations from these semantic priors, inducing a structural shift in how models process rules.

\subsection{Setup: Probing Prior-Context Conflict}

We probe semantic inertia through next-token prediction on 45 scenarios from Tier-1 and Tier-2, focusing on states $S$ with counterintuitive rules (\eg, \texttt{WALL IS YOU}). Models predict the next viable move under two representational modalities:
\textbf{Natural Language:} Descriptive prompts like ``The Wall is You. Valid moves are~...''
\textbf{Code Grounding:} Explicit transition functions $T(\cdot; R_t)$ parameterized by the active rule-set, where $R_t$ specifies transformations such as \texttt{WALL} $\rightarrow$ \texttt{YOU}.
We quantify adaptation through $\Delta P$, the probability gap between logic-driven and prior-driven predictions:
\begin{equation}
    \Delta P = P(w_{\text{logic}}\,|\,S) - P(w_{\text{prior}}\,|\,S).
\end{equation}
Negative values indicate semantic inertia---defaulting to learned associations over contextual logic. We evaluate three model families (Llama-3, Pythia, Qwen2.5) spanning 160M to 70B parameters (\cref{sec:supp:exp_1}).

\begin{figure}[ht!]
    \centering
    \includegraphics[width=\linewidth]{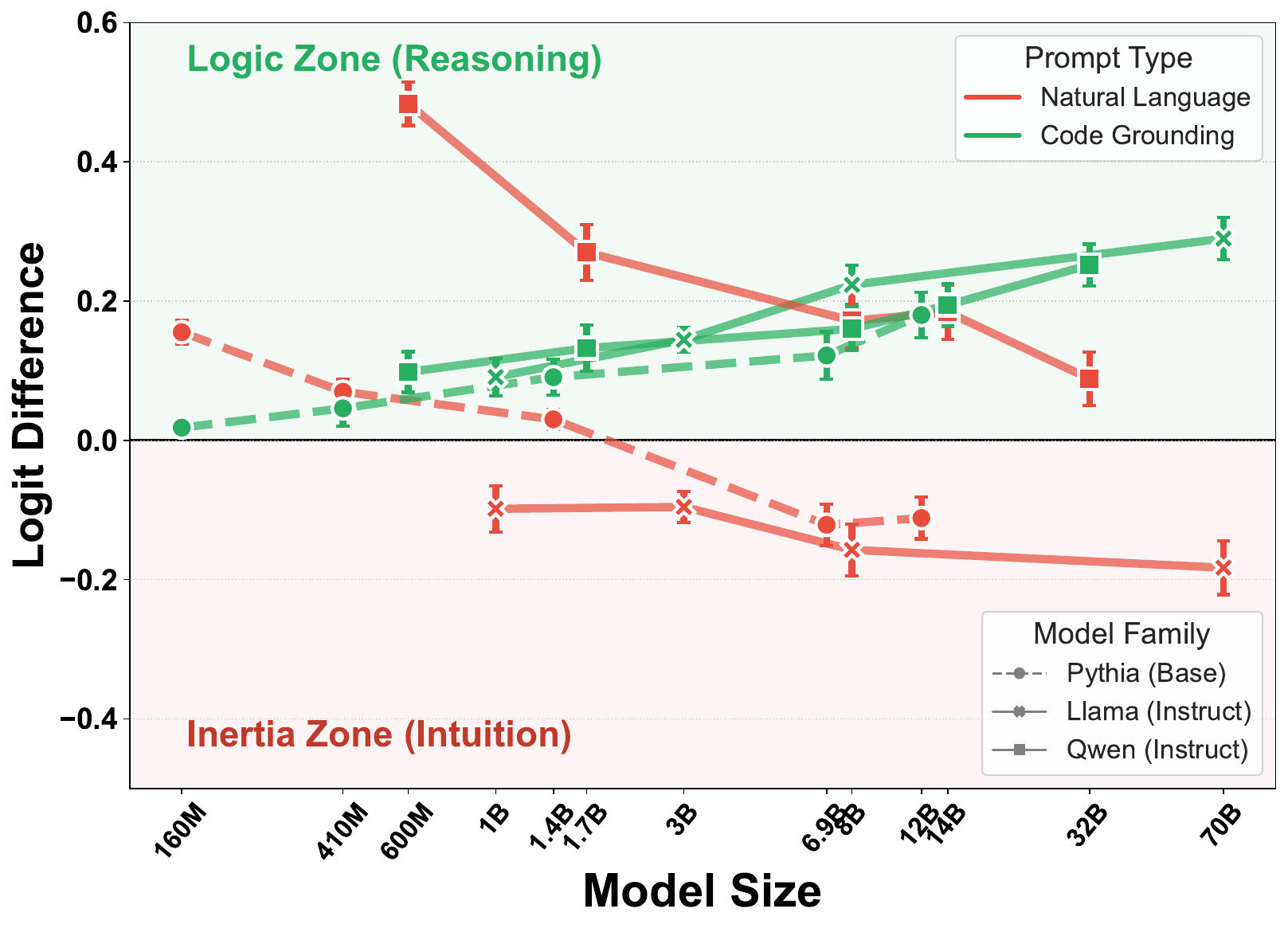}
    \caption{\textbf{Inverse scaling in natural language \vs restored scaling in code.} In Natural Language prompting (red lines), larger models often exhibit worse performance due to semantic interference, with models showing clear inverse scaling. Code Grounding (green lines) decouples logical operations from semantic priors, restoring positive scaling laws where computational scale translates to improved contextual reasoning.}
    \label{fig:scaling_trends}
\end{figure}

\subsection{Results: Code Reverses Inverse Scaling}

\cref{fig:scaling_trends,tab:prob_diff_results} reveal divergent scaling trajectories across representational modalities. Natural language prompting (red curves) exhibits inverse scaling: Llama-3-70B shows stronger semantic inertia ($\Delta P = -0.18$) than its 8B counterpart. Rather than improving with scale, larger models become more entrenched in distributional priors, resisting counterintuitive rule updates.

Code grounding (green curves) inverts this pattern. By expressing game physics as variable assignments, we strip symbols of their semantic associations, shifting computation from pattern matching to logical inference. Under this representation, scale becomes beneficial: Llama-3-70B achieves $\Delta P = +0.29$, a $0.47$ improvement over its natural language performance.

This dissociation reveals that failures in ontological restructuring stem from \textit{representational interference}, not reasoning limitations. Natural language activates distributional priors that override contextual logic; code suppresses these priors, allowing models to leverage their capacity for rule-based reasoning. This finding directly motivates our \ac{lcv} framework, which systematically exploits code's semantic-filtering properties.

\section{\texorpdfstring{\ac{lcv}}{} via Counterfactual Alignment}\label{sec:method}

We address semantic inertia by replacing natural-language action prediction with executable world modeling. Rather than learning a direct policy $\pi_\theta(a_t \mid s_t)$---which conflates perception with dynamics and defaults to pre-trained priors---we learn an amortized theory inducer $f_\theta$ that compiles state and rule-set into a Python transition kernel: $\hat{T}_t \leftarrow f_\theta(s_t)$. A classical planner then searches over actions using $\hat{T}_t$ as its dynamics oracle, re-synthesizing the kernel whenever rules change.

\begin{figure*}[t!]
    \centering
    \includegraphics[width=\linewidth]{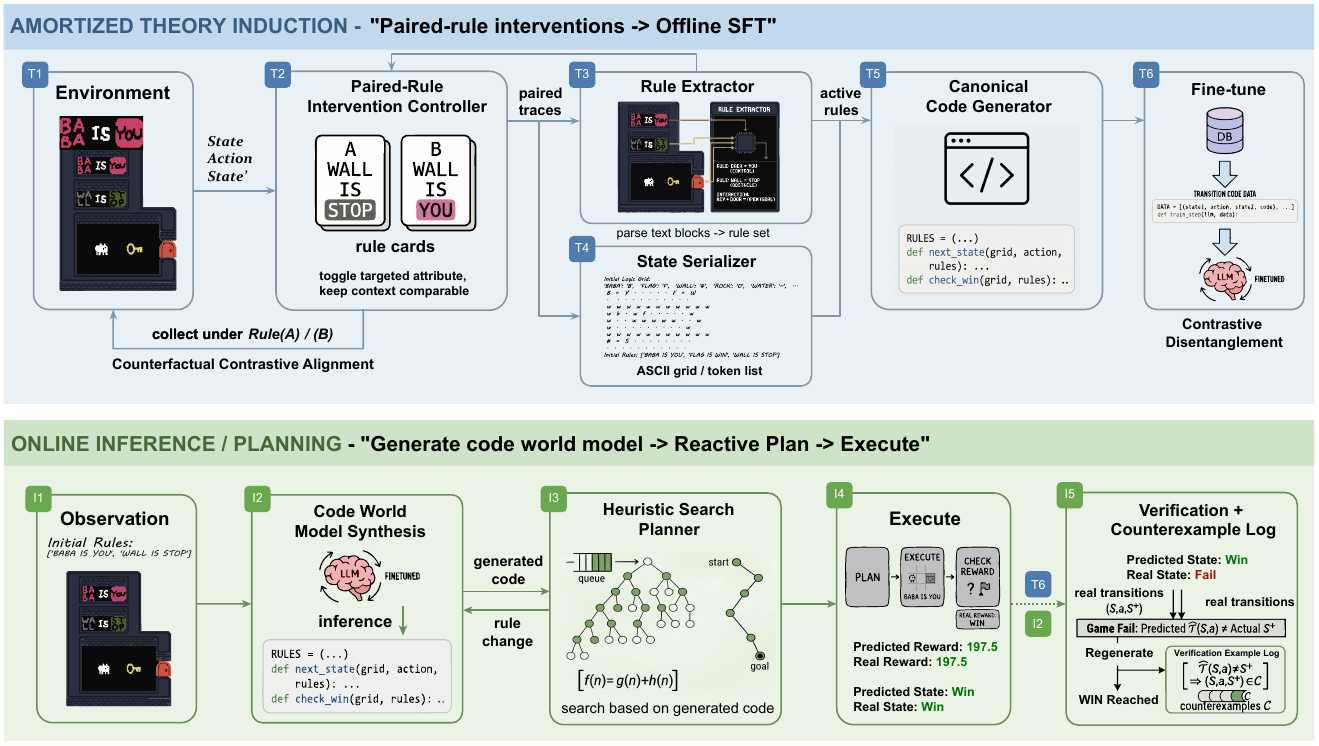}
    \caption{\textbf{Overview of \ac{lcv}.} (Up) \textbf{Amortized Contrastive Theory Induction}: We perform \ac{sft} on paired samples with the same environment state $s$ but contradictory rule-sets $r$ (\eg, \texttt{WALL IS STOP} \vs \texttt{WALL IS YOU}. This directly targets \textit{semantic inertia} by making surface symbols non-diagnostic: the model must generate different executable kernels solely from the active rules, disentangling object names from their pre-trained affordance priors. (Down) \textbf{Online Inference and Planning}: At test time, the model synthesizes the executable transition theory in a single pass, which is compiled and used by a classical planner.}
    \label{fig:baba_pipeline}
\end{figure*}

To ensure $\hat{T}_t$ depends on mutable rules $R_t$ rather than visual semantics, we train with counterfactual contrastive alignment: paired examples share identical grids but contradictory rule-sets, forcing the model to generate different code for the same visuals. After \ac{sft}, we run a reactive loop that re-synthesizes the world model and exploits its efficiency for heuristic search over actions and rule configurations (\cref{fig:baba_pipeline}).

\subsection{Amortized Theory Induction}

Recent neuro-symbolic approaches like TheoryCoder \citep{ahmed2025synthesizing} rely on inference-time search to discover valid programs---a computationally expensive process. We instead amortize reasoning by learning $f_\theta : (s_t) \rightarrow C$, mapping states to executable Python programs representing local physics.

This shift from search to induction provides two benefits. First, efficiency: inference becomes $\mathcal{O}(1)$ (single forward pass) versus $\mathcal{O}(N)$ (iterative debugging). Second, robustness: direct theory prediction bypasses local minima in search-based methods, where internal verification heuristics may reject counterintuitive rules due to prior bias.

\subsection{Counterfactual Contrastive Alignment}

Standard \ac{sft} fails to overcome semantic inertia because it does not penalize reliance on pre-trained priors. When visual state $v$ strongly implies specific dynamics (\eg, ``Wall'' sprite $\rightarrow$ ``Stop''), models minimize loss by attending to visual features while ignoring explicit rules $r$.

To compel the model to ground its reasoning exclusively in the logical rules, we employ a counterfactual sampling strategy to construct a paired training corpora $\mathcal{D}_{\text{pair}}$. The goal is to render the visual signal $v$ ambiguous with respect to the output. For a fixed map configuration $v$, we sample paired rule-sets $r^+$ and $r^-$ that describe contradictory physics, along with their corresponding ground-truth programs $c^+$ and $c^-$:
\begin{equation}
    \begin{aligned}
        \text{Instance A:} \quad & \langle v, r^+ = \text{\texttt{WALL IS STOP}}, c^+ \rangle \\
        \text{Instance B:} \quad & \langle v, r^- = \text{\texttt{WALL IS PASS}}, c^- \rangle
    \end{aligned}
\end{equation}

We employ contrastive disentanglement objective $\mathcal{L}_{\text{CD}}$, treating rule-set $r$ as causal driver and visual prior $v$ as distractor. The objective maximizes joint likelihood of paired codes conditioned on their specific rules:
\begin{equation}
    \begin{aligned}
        \mathcal{L}_{\text{CD}}(\theta)
        = \mathbb{E}_{\mathcal{D}_{\text{pair}}}\Big[
        &- \log P_\theta(c^+ \mid v, r^+) \\
        &- \lambda \log P_\theta(c^- \mid v, r^-)
        \Big]
    \end{aligned}
\end{equation}
We set $\lambda=2$ to emphasize counterfactual learning. While formulated as a joint likelihood rather than a metric loss (\eg, InfoNCE), $\mathcal{L}_{\text{CD}}$ functions as an implicit contrastive mechanism: since $v$ remains invariant while $c^+$ and $c^-$ contradict each other, gradients relying solely on visual priors oscillate and cancel. To minimize $\mathcal{L}_{\text{CD}}$, the model must suppress ambiguous signal in $v$ and attend exclusively to differentiating variable $r$, orthogonalizing logical dynamics from visual appearance.

\subsection{Reactive Planning with Synthesized Vistas}

Amortized induction enables low-latency inference: small models without iterative revision support fully reactive planning loops. In \game, non-stationary transition function $T$ changes when agents rearrange text blocks, requiring frequent world model re-synthesis.

At timestep $t$, our \ac{lcv} executes: (i) fine-tuned model predicts Python class $\hat{T}_t = f_\theta(s_t)$---unlike search-based methods requiring seconds for verification, our $\mathcal{O}(1)$ inference enables instant re-synthesis when rule-state $R_t$ changes; (ii) $\hat{T}_t$ compiles into executable \texttt{next\_state} function; (iii) bounded Greedy Best-First Search uses $\hat{T}_t$ as transition oracle, with domain-agnostic heuristic $h(s)$ prioritizing states reducing Manhattan distance to active \texttt{WIN} objects or interactable text blocks. This reactive re-compilation handles ``Tier 3'' scenarios without pre-defined planners like \ac{pddl}.

\section{Experiments and Results}

\begin{table*}[ht!]
    \centering
    \small
    \caption{\textbf{Success Rate (SR) on \benchmark.} Top: four open-source base models, reported SR only and arranged two models side by side. Bottom: for Deepseek, TheoryCoder, proprietary references, and our LCV models, we additionally report the number of thinking and answer tokens (Tok: Think / Ans) per tier when available.}
    \label{tab:main_results}
    \renewcommand{\arraystretch}{1.2}
    \setlength{\tabcolsep}{6pt}
    \resizebox{\linewidth}{!}{%
        \begin{tabular}{
                l
                c c c
                @{\hspace{1.2cm}}
                l
                c c c
            }
            \toprule
               SR\,$\uparrow$ & \textbf{Tier 1} & \textbf{Tier 2}  & \textbf{Tier 3}  
             & & \textbf{Tier 1}   & \textbf{Tier 2} & \textbf{Tier 3}  \\
             \cmidrule(r){1-4}
            \cmidrule(l){5-8}
            \textbf{\textit{Qwen2.5-7B-Instruct}} & & & & \textbf{\textit{Qwen2.5-72B-Instruct}}&&&\\
            Direct Policy
              & 8.89\%  & 2.22\%  & 0.00\%
              & Direct Policy
              & 13.33\% & 6.67\%  & 2.22\% \\
            Direct Policy (CoT)        
              & 13.33\% & 6.67\%  & 4.00\%
              & Direct Policy (CoT)
              & 28.89\% & 20.00\% & 6.67\% \\
            Code-as-Policy             
              & 13.33\% & 11.11\% & 4.00\%
              & Code-as-Policy
              & 62.22\% & 53.33\% & 42.00\% \\
            \cmidrule(r){1-4}
            \cmidrule(l){5-8}
            \textbf{\textit{GPT-OSS-120B}} & & & &\textbf{\textit{Llama-3-70B-Instruct}} & & & \\
            Direct Policy
              & 8.89\% & 2.22\% & 2.00\%
              & Direct Policy
              & 15.56\% & 6.67\% & 8.00\% \\
            Direct Policy (CoT)        
              & 8.89\% & 2.22\% & 4.00\%
              & Direct Policy (CoT)
              & 17.78\% & 6.67\% & 8.00\% \\
            Code-as-Policy             
              & 6.67\% & 6.67\% & 8.00\%
              & Code-as-Policy
              & 53.33\% & 26.67\% & 18.00\% \\
            \midrule\midrule
        \end{tabular}
    }
    \setlength{\tabcolsep}{6pt}
    \resizebox{\linewidth}{!}{%
        \begin{tabular}{
            l
            c c    
            c c    
            c c    
        }
        \multicolumn{1}{c}{} 
            & \multicolumn{2}{c}{\textbf{Tier 1}} 
            & \multicolumn{2}{c}{\textbf{Tier 2}} 
            & \multicolumn{2}{c}{\textbf{Tier 3}} \\
        \cmidrule(r){2-3}
        \cmidrule(r){4-5}
        \cmidrule(r){6-7}
          & SR\,$\uparrow$ & Tok (Think / Ans)
          & SR\,$\uparrow$ & Tok (Think / Ans)
          & SR\,$\uparrow$ & Tok (Think / Ans)\footnotemark  \\
        \midrule
        \multicolumn{7}{l}{\textbf{\textit{Deepseek-v3.2}}} \\
        Direct Policy      
          & 17.78\% & 7.38k / 15.2
          & 13.33\% & 16.0k\footnotemark / 12.8
          & 8.00\%  & 16.0k / 18.5 \\
        Code-as-Policy                   
          & 40.00\% & 4.24k / 1.16k
          & 31.11\% & 5.67k / 892.4
          & 36.00\% & 3.93k / 962.9 \\
        \midrule
        \multicolumn{7}{l}{\textbf{\textit{Claude Sonnet 4.5}}} \\
        Direct Policy      
          & 57.78\% & 1.24k / 14.7
          & 13.33\% & 3.31k / 11.3
          & 8.00\%  & 3.15k / 16.9 \\
        Code-as-Policy                   
          & 68.89\% & 2.52k / 834.2
          & 31.11\% & 2.87k / 943.6
          & 30.00\% & 2.22k / 677.4 \\
        \midrule
        \multicolumn{7}{l}{\textbf{\textit{Gemini 3 Pro Preview}}} \\
        Direct Policy      
          & 62.22\% & 1.43k / 13.5
          & 13.33\% & 2.07k / 19.1
          & 10.00\%  & 2.00k / 10.4 \\
        Code-as-Policy                   
          & 71.11\% & 1.27k / 953.4
          & 31.11\% & 1.65k / 1.05k
          & 28.00\% & 1.18k / 1.14k \\
        \midrule
        
        \textbf{\textit{TheoryCoder}}\\
        GPT-4o, 1 API call
          & 24.44\% & -- / 1.13k 
          & 13.33\% & -- / 1.31k
          & 8.00\%  & -- / 1.31k \\
        GPT-4o           
          & 62.22\% & -- / 2.46k 
          & 53.33\% & -- / 3.22k
          & 52.00\% & -- / 3.08k \\
        \midrule
        \textbf{Human (with tutorial video)} 
          & \underline{95.56\%} & - 
          & \underline{82.22\%} & - 
          & \underline{78.00\%} & - \\
        \midrule
        \multicolumn{7}{l}{\textbf{\textit{Ours (Amortized LCV)}}} \\
        \textbf{LCV (Vanilla $L_{\mathrm{SFT}}$)}    
          & 88.89\% & -- / 763.9 
          & 60.00\% & -- / 1.25k 
          & 48.00\% & -- / 1.27k \\
        \textbf{LCV (Contrastive $L_{\mathrm{CD}}$)} 
          & \textbf{93.33\%} & -- / 798.5 
          & \textbf{75.56\%} & -- / 1.16k 
          & \textbf{62.00\%} & -- / 1.31k \\
        \bottomrule
        \end{tabular}%
    }%
\end{table*}

\footnotetext[2]{For Tier 3, since the rules can change, we report only the average token counts for each plan.}
\footnotetext[3]{16k indicates that the maximum generation token limit was reached.}

We evaluate the efficacy of \ac{lcv} in mitigating semantic inertia and enabling robust planning under dynamic ontologies. We try to answer three questions:

\noindent\textbf{RQ1 (Performance Hierarchy):} Does decoupling theory induction (code synthesis) from planning outperform end-to-end neural policy generation, particularly in semantically adversarial settings?

\noindent\textbf{RQ2 (Inhibitory Control):} To what extent does the counterfactual contrastive alignment objective reduce prior-reversion behavior compared to standard supervision?

\noindent\textbf{RQ3 (Amortization Efficiency):} While a fine-tuned small model (7B) can surely outperform inference-heavy approaches (\eg, TheoryCoder with GPT-4o) in latency and robustness, how do these gains trade off against the cost of data collection and \ac{sft}?

\subsection{Baselines} 
We compare our \ac{lcv} against a diverse set of neural and neuro-symbolic baselines: 

\noindent \textbf{Direct Policy (Zero-Shot / CoT):} Standard agentic prompting where the model predicts actions directly. We evaluate both standard IO and Chain-of-Thought (CoT) \citep{wei2022chain}.

\noindent \textbf{Code-as-Policy (CaP):} Following \citet{liang2023code}, the model generates a heuristic Python script to solve the task directly, without explicitly modeling the transition dynamics.

\noindent \textbf{TheoryCoder:} We implement the inference-time variant of \citet{ahmed2025synthesizing}, in which GPT-4o iteratively synthesizes a transition function through few-shot prompting and execution feedback. While recent baselines such as Chain of Code \citep{li2024chain} also address code-centric reasoning, we believe TheoryCoder targets a closely related class of problems and represents a state-of-the-art approach for the coding–planning pipeline.

We evaluate open-weight models (Qwen, LLaMA, DeepSeek, GPT-OSS) and proprietary foundations (Gemini 3 Pro, Claude Sonnet 4.5, GPT-4o).  Prompts and detailed settings can be found in \cref{sec:supp:prompt}.

For our \ac{lcv} agent, we fine-tune a small Qwen2.5-7B-Instruct model using approximately 600 paired training samples collected from \benchmark for world model synthesis. The synthesized world model is compiled and passed to a bounded heuristic planner with a node expansion budget of $N{=}2000$. We report the average Success Rate (SR) together with \acs{llm}-generated token lengths (ToK) for each tier. For evaluation, we sample 45 paired environments for both Tier 1 and Tier 2, and 50 environments for Tier 3.

\subsection{Main Results}

\cref{tab:main_results} summarizes the performance across the three complexity tiers. Also, solution examples can be found in \cref{sec:supp:solution_example}.

\noindent \textbf{The Collapse of Natural Language (RQ1).} 
Standard prompting reveals a catastrophic failure of inhibition. While foundation models like Claude 4.5 Sonnet perform strongly on aligned tasks (Tier 1: 57.78\%), they collapse on adversarial tasks (Tier 2: 13.33\%). Notably, huge models like GPT-OSS-120B and Llama-3-70B show negligible improvement over their smaller counterparts in Tier 2. This empirically confirms our \textit{Inverse Scaling} hypothesis: scale alone does not solve functional fixedness. The visual prior of a ``Wall'' overrides the logical instruction ``Wall is Pass,'' leading to persistent hallucination of obstacles regardless of model size.

\noindent \textbf{Code as a Scaffold, Logic as a Solution.} 
Methods utilizing code (CaP, TheoryCoder) show greater resilience. However, \ac{lcv} achieves dominant performance, with 75.56\% SR in Tier 2, significantly outperforming other baselines. This result highlights that simply using code (CaP) is insufficient if the generation process itself is biased by priors. By separating the generation of physics from the generation of actions, \ac{lcv} isolates the reasoning error, allowing the planner to search a clean, logic-grounded state space.

\noindent \textbf{Robustness in Dynamic Settings (Tier 3).} 
Tier 3 requires manipulating rules to change the environment mid-trajectory. Here, TheoryCoder's performance plateaus, likely due to error accumulation in its iterative synthesis loop. In contrast, our amortized approach maintains robustness. Because our model is trained to map state-configurations to code directly, it treats a mid-episode rule change simply as a new input state, re-compiling the physics engine instantly without the need for fragile conversational history maintenance.

\subsection{Analysis of Inhibitory Control (RQ2)}

To isolate the mechanism of improvement, we explicitly compare \ac{lcv} (Vanilla) \vs \ac{lcv} (Contrastive) in \cref{tab:main_results}. In normal cases, the performance gap is marginal ($88.9\%$ vs $93.3\%$). Both models easily learn the mapping when visuals and logic agree. In Tier 2, the gap widens significantly ($60.0\%$ vs $75.6\%$). This differential gain validates our gradient orthogonalization hypothesis (Section \ref{sec:method}). The Vanilla model, trained with standard \ac{sft}, still overfits to spurious correlations (\eg, associating Wall sprites with `stop=True`). The paired contrastive objective forces the model to attend to the rule-text to distinguish between the paired samples, thereby embedding the inhibitory control directly into the model weights.

\subsection{Ablation 1: Learning Efficiency}

A dominant paradigm in recent reasoning literature is the scaling of inference-time compute---using iterative scaffolding (\eg, CoT, TheoryCoder loops) to resolve ambiguity \citep{shinn2023reflexion}. Our results challenge the universality of this approach in adversarial settings. As shown in \cref{tab:efficiency}, approaches like TheoryCoder (GPT-4o) require extensive token expenditure ($\sim$3.2k/problem) to ``debug'' their way out of semantic priors. We argue that this represents an inefficient allocation of compute: the model is burning tokens to fight its own parametric memory. By shifting this burden to training-time alignment, \ac{lcv} achieves a $\mathbf{4\times}$ reduction in latency while improving accuracy. This suggests that inhibitory control is amortizable. The cognitive effort required to dissociate ``Wall'' from ``Stop'' does not necessarily require explicit step-by-step reasoning at every instance. Through our contrastive alignment, we effectively compile this inhibition into the model's forward pass, transforming a complex reasoning task into a rapid, reflex-based retrieval. This offers a scalable path for deploying robust agents in real-time environments where iterative ``thinking'' loops are prohibitively slow.

\subsection{Ablation 2: Generalization}

A frequent critique of fine-tuning small models (7B) versus prompting large foundations (72B+) is the potential loss of generalization. We probe this via two distinct splits (detailed in \cref{sec:supp:generalization}):

\noindent \textbf{Map Generalization (Spatial Robustness).} 
On unseen spatial configurations, \ac{lcv} retains almost all of its in-distribution performance ($76.4\% \to 74.3\%$). This result is pivotal: it indicates that our model has not overfitted to visual patterns (\eg, ``Wall at pixel $x,y$''). Instead, it has learned a symbolic compiler which can be seen as a function that maps visual discrete entities to abstract logical rules, regardless of their spatial distribution. 

\noindent \textbf{Combination Generalization (Logical Robustness).} 
The ``Combo Gen'' split (\eg, introducing \texttt{WALL IS MELT}) requires systematicity---combining known atoms in novel ways. While all models degrade, \ac{lcv} maintains a massive lead over the baselines ($72.4\%$ \vs $48.2\%$ for Qwen-72B). 

In conclusion, standard \acp{llm} rely on parametric generalization---solving novel tasks by mapping them to similar examples in the pre-training corpus. When the task explicitly contradicts that corpus (as in Tier 2/3), parametric generalization becomes a liability. \ac{lcv} exhibits systematic generalization: because it operates on an intermediate code representation, it can compose rules it has never seen together, provided the underlying syntax (Python Code) remains consistent. This confirms that code-grounding acts as a regularizer, forcing the model to learn the grammar of physics rather than the statistics of scenarios.

\section{Conclusion}

In this work, we demonstrate that semantic inertia scales inversely with model size, hampering robust reasoning in dynamic environments. We propose \acf{lcv}, a framework that decouples logical dynamics from visual priors via amortized theory induction and counterfactual contrastive alignment. Our approach significantly outperforms inference-heavy baselines like TheoryCoder in both robustness and efficiency. These findings suggest that for mutable ontologies, reasoning must be grounded in executable, verifiable code rather than probabilistic natural language.

\section*{Limitations}

While \ac{lcv} effectively mitigates semantic inertia, it relies on discrete state abstractions (grid worlds) and may not directly transfer to continuous, high-dimensional visual domains without an additional perception module. Furthermore, our Counterfactual Contrastive Alignment requires paired data with contradictory rules, which is procedurally generatable in logic puzzles but potentially expensive to curate in naturalistic environments.

\section*{Acknowledgements}

This work is supported in part by the National Natural Science Foundation of China (32595491,62376009), the State Key Lab of General AI at Peking University, the PKU-BingJi Joint Laboratory for Artificial Intelligence, the Wuhan Major Scientific and Technological Special Program (2025060902020304), the Hubei Embodied Intelligence Foundation Model Research and Development Program, and the National Comprehensive Experimental Base for Governance of Intelligent Society, Wuhan East Lake High-Tech Development Zone.

\bibliography{reference_header,reference}
\clearpage

\appendix
\renewcommand\thefigure{A\arabic{figure}}
\setcounter{figure}{0}
\renewcommand\thetable{A\arabic{table}}
\setcounter{table}{0}
\renewcommand\theequation{A\arabic{equation}}
\setcounter{equation}{0}
\pagenumbering{arabic}
\renewcommand*{\thepage}{A\arabic{page}}
\setcounter{footnote}{0}

\section{Environment Overview: \game}

\begin{figure}[t!]
    \centering
    \includegraphics[width=\linewidth]{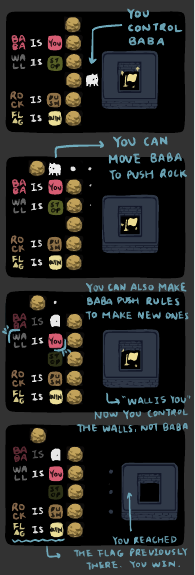}
    \caption{Overview of the \game environment. }
    \label{fig:supp:game_overview}
\end{figure}

\game\footnote{https://hempuli.com/baba/} is an award-winning logic puzzle game developed by Arvi Teikari (Hempuli) that fundamentally reimagines the mechanics of grid-based interactions. Unlike traditional environments (\eg, Sokoban or Minecraft) where object affordances are static constants---a wall is always an obstacle, a key is always an item---\game treats the rules of physics as tangible, manipulatable objects within the game world. The environment consists of ``Word'' blocks (\eg, \texttt{WALL}, \texttt{IS}, \texttt{YOU}) and physical objects (sprites). A rule is formed only when three Word blocks align syntactically (Noun-Operator-Property). Crucially, the player can push these Word blocks to rewrite the game's logic in real-time. For instance, disconnecting \texttt{WALL IS STOP} instantly renders walls permeable, while forming \texttt{ROCK IS YOU} transfers the player's agency from the titular character Baba to a Rock. This unique mechanism collapses the distinction between the object level and the meta-level, creating a testbed where an agent cannot rely on fixed semantic priors (\eg, ``skulls are dangerous'') but must actively ground its reasoning in the current, mutable configuration of the text blocks.

\section{Inverse Scaling Details}\label{sec:supp:exp_1}

\paragraph{Scenario Construction.}
We curated 45 scenarios from the \game\ environment, specifically targeting states where the current rule set induces a sharp conflict with typical semantic priors (e.g., \texttt{LAVA IS SAFE}, \texttt{WALL IS YOU}). For each scenario, the board state and active rules were manually defined to create pairs of logic-driven and prior-driven actions. Each input includes the grid, a list of possible moves, and the set of rules currently in effect.

\paragraph{Prompt Engineering.}
For each scenario, two types of prompts were prepared:
\begin{itemize}
    \item \textbf{Natural Language (NL) Prompts:} These describe the current environment and explicitly state the rules in plain language, e.g., “The rule is: 'Wall is You'. Valid moves are: UP, DOWN, LEFT, RIGHT. Which is correct under the current rules?”
    \item \textbf{Code Grounding Prompts:} These supply the rule set as code-like statements or as arguments to an explicit transition function, emphasizing variable assignment and logic, e.g., “Given: rules = {'Wall': 'You', ...}. Which action is valid according to the transition function $T(\cdot; R_t)$?”
\end{itemize}
To avoid position biases, the order of presenting moves and states was randomized across prompts.

\paragraph{Model Families and Parameter Scaling.}
Evaluation was performed using freely available and open-source large language models: Llama-3 (1B, 8B, 70B), Pythia (160M, 1.4B, 12B), and Qwen2.5 (600M, 8B, 32B). Both base and instruction-tuned checkpoints were included where compatible with the interface. All model inference was conducted using \texttt{transformers} with greedy decoding (temperature 0), and where possible, probabilities for all candidate single-token actions (UP, DOWN, LEFT, RIGHT) were directly obtained from output logits.

\paragraph{Inference Setup.}
Models were run on a single NVIDIA A100 node. Each prompt was independently batched and sent to the model, and output log probabilities for each candidate action were retrieved. All models are deployed locally.

\paragraph{Labeling and Score Computation.}
For each prompt, we annotated which action is logic-driven (i.e., required by the current non-default rules) and which action conforms to the most probable prior (e.g., path avoidance of 'dangerous' lava or 'impassable' wall). We then computed the adaptation score for each instance as:
\[
\Delta P = P(w_{\text{logic}}\,|\,S) - P(w_{\text{prior}}\,|\,S)
\]
Average $\Delta P$ scores were calculated separately for Natural Language and Code Grounding modalities, across all parameter sizes within each family.

\paragraph{Additional Notes.}
To ensure that models were not overfitting to specific lexical cues, several variants of the scenarios were repeated with shuffled agent/object identities (e.g., swapping 'BABA' and 'ROCK'), confirming that trends in inverse scaling were robust to superficial changes. In all cases, the main observation held: larger models in the NL condition exhibited stronger semantic inertia, whereas code grounding robustly enabled scaling benefits for context-driven reasoning.

\section{\texorpdfstring{\ac{lcv}}{} Implementation Details}

\subsection{\acs{sft}}

For our main experiments, we fine-tuned the Qwen2.5-7B-Instruct model, which has 7 billion parameters, using the LLaMA-Factory framework and LoRA adaptation. Training was performed for 20 epochs with a batch size of 1 per device (gradient accumulation steps: 4) on 4 NVIDIA RTX 3090 GPUs (24GB each), using FP16 precision. The total GPU compute for fine-tuning was approximately 10 GPU-hours. In all experiments, we applied LoRA ($r=8$, $\alpha=16$, dropout $=0$) on Qwen2.5-7B-Instruct, targeting all (\texttt{k\_proj}, \texttt{q\_proj}, \texttt{v\_proj}, \texttt{o\_proj}, \texttt{up\_proj}, \texttt{down\_proj}, \texttt{gate\_proj}) projection layers, without bias or additional PEFT techniques. All hyperparameters are provided to facilitate reproducibility.

\subsection{Overall Planning Pipeline}

In each planning episode, the search operates over an explicit symbolic graph, where nodes correspond to compressed grid states $s \in \mathcal{S}$ and edges correspond to agent actions $a \in \mathcal{A}$. Unlike environments with static transition dynamics, \game defines state transitions and terminal conditions through a logic model $\mathcal{M}$ that is generated at runtime by the \ac{llm}. Crucially, the system does not hard-code notions such as controllability, victory, or physical affordances. Instead, it queries the current state and the active rule set to infer all relevant entities and interactions.

At every node expansion, the search consults $\mathcal{M}$ to parse symbolic rules expressed as triples of the form
\[
(\text{Subject},\ \texttt{IS},\ \text{Property}),
\]
such as \texttt{BABA IS YOU}, \texttt{FLAG IS WIN}, or \texttt{WALL IS STOP}. Based on the currently active rules in state $s$, the planner dynamically constructs entity sets associated with each property. For example:
\begin{align}
    \mathcal{P}_{\text{you}}  &= \{ p : p \text{ satisfies } (\cdot,\texttt{IS},\texttt{YOU}) \text{ in } s \}, \\
    \mathcal{P}_{\text{win}}  &= \{ p : p \text{ satisfies } (\cdot,\texttt{IS},\texttt{WIN}) \text{ in } s \}, \\
    \mathcal{P}_{\text{push}} &= \{ p : p \text{ satisfies } (\cdot,\texttt{IS},\texttt{PUSH}) \text{ in } s \}, \\
    \mathcal{P}_{\text{stop}} &= \{ p : p \text{ satisfies } (\cdot,\texttt{IS},\texttt{STOP}) \text{ in } s \}.
\end{align}
These sets may change dynamically as agent actions modify the map‘s logic, for instance by rearranging textual blocks $\mathcal{T}$ to create or destroy rules such as \texttt{ROCK IS PUSH} or \texttt{WALL IS STOP}.

Planning is guided by logic-aware heuristics that adapt to the currently active rule configuration. When a winning condition exists (\eg, \texttt{FLAG IS WIN}), the heuristic encourages controllable entities to approach winning objects:
\[
    h(s) = \min_{p_i \in \mathcal{P}_{\text{you}},\; p_j \in \mathcal{P}_{\text{win}}}
    \left( |x_i - x_j| + |y_i - y_j| \right).
\]
If no win condition is present, or if the current rules introduce obstacles or hazards (\eg, \texttt{WALL IS STOP} or \texttt{LAVA IS DEFEAT}), the heuristic instead prioritizes interactions that may alter the rule set. In such cases, the search is biased toward textual elements:
\[
    h(s) = \min_{p_i \in \mathcal{P}_{\text{you}},\; t \in \mathcal{T}}
    \left( |x_i - x_t| + |y_i - y_t| \right).
\]

Beyond simple navigation, the planner reasons about more complex emergent behaviors induced by rule composition. For instance, if \texttt{LAVA IS DEFEAT} is active, lava tiles are treated as lethal regions to be avoided. If both \texttt{KEY IS OPEN} and \texttt{DOOR IS SHUT} are present, the planner can infer the possibility of unlocking doors through interaction. Importantly, the system does not assume a fixed vocabulary of properties; instead, it parses and reasons over any rule discovered by $\mathcal{M}$, provided it conforms to the \texttt{X IS Y} syntactic structure.

For each action $a \in \mathcal{A}$ taken from state $s$, a candidate successor state $s'$ is produced by invoking $\mathcal{M}.next\_state\_fn(s, a)$, and is subsequently interpreted under the newly inferred rule set. Cycle detection via state hashing prevents redundant exploration, while a priority queue enables best-first expansion according to the current logic-aware heuristic.

The planning process terminates when a state is reached in which any controllable entity satisfies an active winning rule (i.e., a \texttt{($\cdot$\ IS\ WIN)} triple applies), or when predefined computational resource limits are exceeded.

\begin{algorithm}[htbp]
    \caption{Symbolic Dynamic Reasoning}
    \Input{Start state $s_0$, WorldModel $\mathcal{M}$, Max Depth $D$}
    
    Initialize $\mathit{OpenSet}$ as a priority queue ordered by $g+h$\;
    Initialize $\mathit{Visited}$ as an empty set\;
    $\mathit{OpenSet}.\text{push}(\text{Node}(s_0, 0, h(s_0), [\,]))$\;
    
    \While{$\mathit{OpenSet}$ is not empty}{
        $n \gets \mathit{OpenSet}.\text{pop}()$\;
        
        \If{$\mathcal{M}.\texttt{check\_win}(n.\text{state})$}{
            \Return $n.\text{path}$\;
        }
        
        \If{$\text{length}(n.\text{path}) > D$}{
            \Continue\;
        }
        
        \For{each action $a \in \mathcal{A}$}{
            $s' \gets \mathcal{M}.\texttt{next\_state\_fn}(n.\text{state}, a)$\;
            
            \If{$\texttt{Hash}(s') \notin \mathit{Visited}$}{
                Parse active rule set $\mathcal{R}$ in $s'$ via $\mathcal{M}$\;
                
                \For{each property $q \in \{\text{YOU, WIN, PUSH, \dots}\}$}{
                    $\mathcal{P}_q \gets \{p : (\cdot, \mathrm{IS}, q) \in \mathcal{R}\}$\;
                }
                
                Compute $h'$ using the updated rule sets\;
                $\mathit{OpenSet}.\text{push}(\text{Node}(s', n.\text{cost} + 1, h', n.\text{path} + [a]))$\;
                $\mathit{Visited}.\text{add}(\texttt{Hash}(s'))$\;
            }
        }
    }
    \Return FAIL\;
\end{algorithm}

This generalized strategy allows the agent to reason flexibly over a diverse and evolving set of symbolic rules, enabling robust planning amid dynamic changes to controllable entities, victory conditions, obstacles, hazards, and interaction affordances.

\subsection{Inference Acceleration under Dynamic Rule Changes}

When the active rules change during planning and require updates to the logic model, the LCV framework must invoke the \ac{llm} to regenerate the executable world model $\mathcal{M}$ conditioned on the new rule configuration. Standard autoregressive generation can be computationally expensive, especially when each update introduces only incremental modifications to the current rules or environment.

To address this, we employ two complementary inference acceleration strategies. First, we leverage the \acs{llm}'s key-value (KV) cache mechanism and integrate TensorRT-based compilation for efficient token generation. Since the underlying structure and output tokens of consecutive world models are often highly similar---particularly when rule changes are local or only affect a subset of the semantic triples---cached computation enables us to skip recomputation of unchanged subtrees and dramatically reduce inference latency. Empirically, these optimizations yield a $4.6\times$ speedup on a single NVIDIA A100 GPU for world model synthesis. 

Second, we implement a rule-based caching scheme at the symbolic logic level. By storing previously synthesized world model code and semantic transitions for encountered rule sets, we avoid redundant regeneration when the agent revisits similar or identical configurations. When a rule change is detected, the framework queries the cache for a matching logic signature; if present, the cached model is reused directly, and only new or altered logic components are synthesized. This cache is implemented both at the level of \ac{llm} prompts and the resulting Python code, and it supports partial rule inheritance to further improve reuse when only local adjustments are required. Empirically, in benchmark experiments, the rule-based cache successfully avoided about 85\% potential world model regenerations.

Together, these optimizations ensure fast turnaround in logic model synthesis, enabling robust and responsive agent planning even as the underlying rule set evolves at runtime.

\section{Mean Adaptation Scores}

\begin{table}[ht!]
    \centering
    \small
    \caption{\textbf{Mean Adaptation Scores ($\Delta P$).} Under Natural Language (NL), larger models often perform worse (Inverse Scaling) due to stronger semantic priors. Code Grounding reverses this, enabling large models (\eg, Llama-3-70B) to effectively inhibit priors and maximize reasoning performance.}
    \label{tab:prob_diff_results}
    \renewcommand{\arraystretch}{1.1}
    \setlength{\tabcolsep}{8pt}
    \begin{tabular}{llcc}
        \toprule
        \textbf{Family} & \textbf{Size} & \textbf{NL} ($\Delta P$) & \textbf{Code} ($\Delta P$) \\
        \midrule
        \multirow{3}{*}{Pythia} 
         & 160M & \textbf{0.155} & 0.018 \\
         & 1.4B & 0.030 & \textbf{0.091} \\
         & 12B & -0.112 & \textbf{0.180} \\
        \midrule
        \multirow{3}{*}{Qwen3} 
         & 0.6B & \textbf{0.484} & 0.098 \\
         & 8B & 0.172 & 0.160 \\
         & 32B & 0.089 & \textbf{0.252} \\
        \midrule
        \multirow{3}{*}{Llama-3} 
         & 1B & -0.098 & \textbf{0.091} \\
         & 8B & -0.157 & \textbf{0.223} \\
         & 70B & -0.183 & \textbf{0.290} \\
        \bottomrule
    \end{tabular}
\end{table}

\section{Prompt Examples}\label{sec:supp:prompt}

All baselines are evaluated under identical task descriptions and prompting protocols, without access to intermediate environment states, external tools, or execution feedback beyond textual observations.

\subsection{Direct Action Plan}

\begin{Verbatim}[breaklines=True]
[Level: 0-0]
You are playing 'Baba Is You'. Win the level.

### Legend
'.': EMPTY, '#': WALL, '=': IS, 'B': BABA, 'F': FLAG, 'O': ROCK, 'P': PUSH, 'S': STOP, 'W': WIN, 'Y': YOU, 'b': ICON_BABA, 'f': ICON_FLAG, 'r': ICON_ROCK, 'w': ICON_WALL

### Active Rules
['BABA IS YOU', 'FLAG IS WIN', 'ROCK IS PUSH', 'WALL IS STOP']

### Map (ASCII)
B = Y . . . . . F = W
. . . . . . . . . . .
. . w w w w w . . . .
. . w . . . w . . . .
. . w . b r f . . . .
. . w . . . w . . . .
. . w w w w w . . . .
. . . . . . . . . . .
# = S . . O = P . . .

### Instructions
1. Identify YOU (the character you control) and WIN (the target).
2. Plan a path. Push text blocks if needed to change rules.
3. Output ONLY the sequence of moves: UP, DOWN, LEFT, RIGHT.

Think step-by-step about what obstacles exist and how to overcome them. (for CoT)
\end{Verbatim}
\subsection{Coding Plan}
\begin{Verbatim}[breaklines=True]
You are a logic engine generator for 'Baba Is You'.
Write Python code to simulate the game step based on the rules.

### Symbol Mapping:
- Objects: 'B'=BABA, 'F'=FLAG, '#'=WALL, 'O'=ROCK, '.'=EMPTY
- Text:    'b','f','w','r' (Small chars are icons/text blocks)
- Logic:   '='=IS, 'Y'=YOU, 'W'=WIN, 'S'=STOP, 'P'=PUSH

### Requirements:
1. **Reference**: Use the skeleton below.
2. **No Numpy**: Use standard python lists.
3. **Logic**: Implement ONLY the active rules listed below.

### Reference Skeleton:
```python

def next_state(grid, action):
    # grid: List[List[str]], \eg, grid[0][0] = 'B' (Baba) or 'Bw' (Baba + Wall)
    # action: str, 'UP', 'DOWN', 'LEFT', 'RIGHT'
    
    rows = len(grid)
    cols = len(grid[0])
    new_grid = [row[:] for row in grid] # Deep copy
    
    # ... Your Logic Here (Move YOU, Handle STOP/PUSH) ...
    
    return new_grid

def check_win(grid):
    # Check if 'YOU' overlaps 'WIN'
    return False

```

### Current Active Rules:
['BABA IS YOU', 'FLAG IS WIN', 'ROCK IS PUSH', 'WALL IS STOP']

### Current Grid (Context):
B = Y . . . . . F = W
. . . . . . . . . . .
. . w w w w w . . . .
. . w . . . w . . . .
. . w . b r f . . . .
. . w . . . w . . . .
. . w w w w w . . . .
. . . . . . . . . . .
# = S . . O = P . . .

### Task:
Output the complete Python code implementing `next_state` and `check_win`.
\end{Verbatim}

\section{Environment Visualization and Complexity Tiers}
\label{sec:supp:env_viz}

\begin{figure*}[t!]
    \centering
    \begin{subfigure}[b]{0.85\linewidth}
        \centering
        \includegraphics[width=\linewidth]{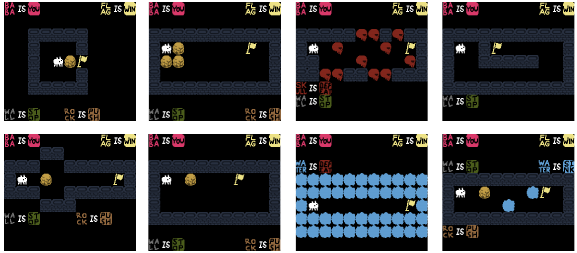}
        \caption{\textbf{Tier 1: Semantic Alignment.} The rules match visual priors (\eg, Walls stop movement), serving as a baseline for spatial planning capability.}
        \label{fig:demo_1}
    \end{subfigure}
    \vspace{0.5cm}
    \begin{subfigure}[b]{0.85\linewidth}
        \centering
        \includegraphics[width=\linewidth]{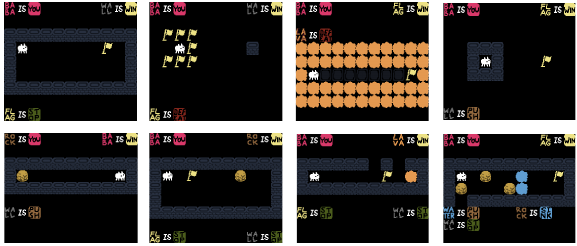}
        \caption{\textbf{Tier 2: Semantic Conflict.} Rules are adversarial to visual intuition. Here, \texttt{LAVA IS SAFE}, requiring the agent to overcome the ``fear'' bias.}
        \label{fig:demo_2}
    \end{subfigure}    
    \vspace{0.5cm}
    \begin{subfigure}[b]{0.85\linewidth}
        \centering
        \includegraphics[width=\linewidth]{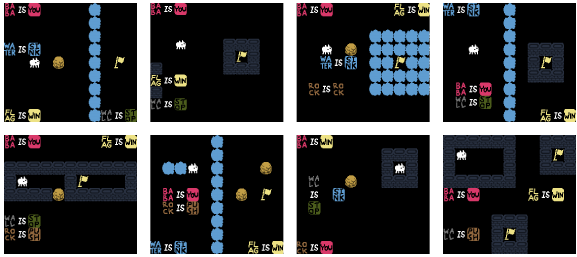}
        \caption{\textbf{Tier 3: Dynamic Plasticity.} The agent must manipulate blocks to rewrite physics (\eg, making walls passable) to solve the puzzle.}
        \label{fig:demo_3}
    \end{subfigure}
    \caption{\textbf{Visualizing the \benchmark Difficulty Hierarchy.} We present three tiers of increasing neuro-symbolic friction. Unlike traditional RL environments with static affordances, our benchmark requires agents to handle: (a) aligned semantics, (b) counterintuitive inhibitory control, and (c) dynamic rule restructuring.}
    \label{fig:bench_tiers_vertical}
\end{figure*}

We provide qualitative visualizations and concrete examples of the three environment tiers used in \benchmark. The tiers are designed to incrementally isolate distinct sources of reasoning failure under dynamic, rule-driven environments. Representative levels from each tier are shown in \cref{fig:bench_tiers_vertical}.

\paragraph{Tier 1: Semantic Alignment (Control).}
Tier 1 environments serve as a control condition for basic spatial reasoning and action sequencing. The active rules are fully aligned with common visual and physical priors learned during pre-training (\eg, \texttt{WALL IS STOP}, \texttt{FLAG IS WIN}). Object appearances and affordances are consistent, allowing agents to rely on standard navigation heuristics. Performance in this tier establishes a baseline and ensures that failures in higher tiers are not attributable to deficiencies in low-level search or perception.

\paragraph{Tier 2: Semantic Conflict (Inhibitory Control).}
Tier 2 environments introduce explicit conflicts between visual semantics and logical rules. Rules such as \texttt{LAVA IS SAFE} or \texttt{WALL IS YOU} deliberately contradict intuitive expectations induced by sprite appearance. Solving these levels requires suppressing pre-trained semantic associations and acting strictly according to the symbolic rule set. This tier isolates failures of semantic inhibition, analogous to Stroop-like interference effects, where visually salient cues must be overridden by abstract task rules.

\paragraph{Tier 3: Dynamic Plasticity (Rule Adaptation).}
Tier 3 environments require multi-stage planning under non-stationary semantics. Levels typically begin in configurations where progress is impossible under the initial rules. Agents must manipulate text blocks to construct new rules (\eg, forming \texttt{WALL IS PASS}) that temporarily alter the environment‘s affordances, and may later need to dismantle or revise these rules to avoid adverse consequences. Success in this tier depends on maintaining an accurate internal representation of the current rule set, updating it after each intervention, and avoiding reliance on outdated semantic assumptions.

Together, these tiers provide a controlled progression from aligned semantics to conflicting and dynamically mutable ontologies, enabling fine-grained analysis of how agents respond to semantic interference and rule-driven concept revision.

\subsection{Solution Sample}\label{sec:supp:solution_example}

\subsection{Gemini Solution Sample}

\begin{figure*}[h!]
    \centering
    \includegraphics[width=\linewidth]{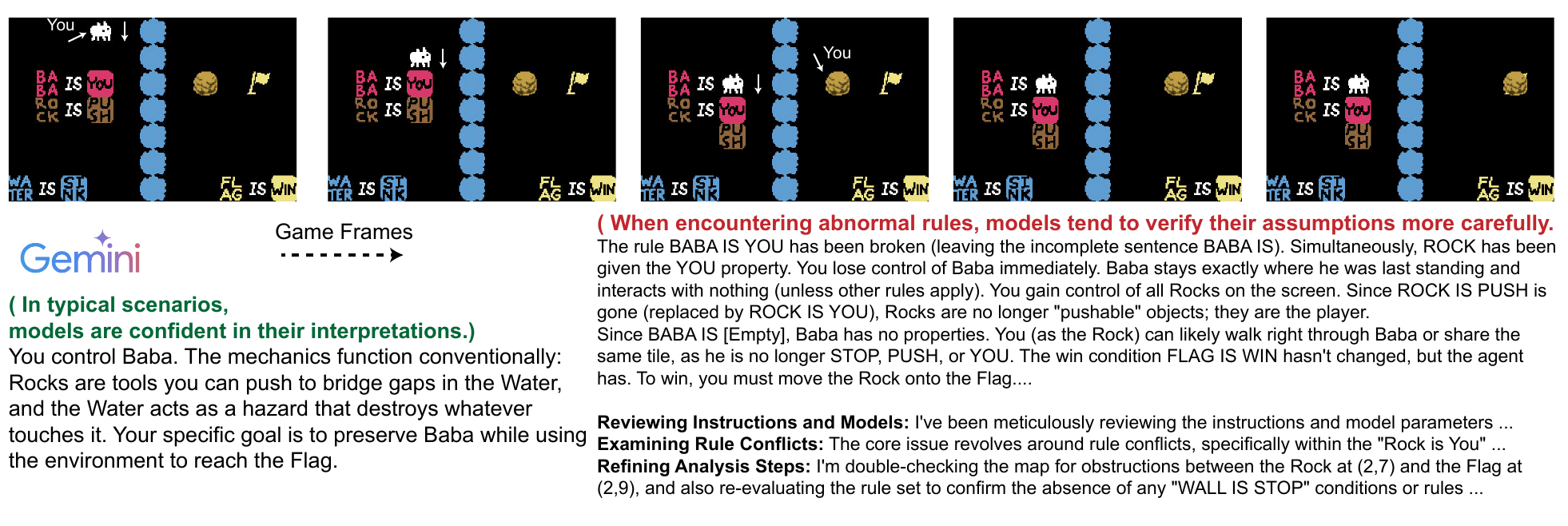}
    \caption{\textbf{Adaptive Inference Utility.} In semantically aligned scenarios (Green), Gemini plans concisely using intuitive priors. However, facing counterintuitive rules like \texttt{ROCK IS YOU} (Red), it spontaneously shifts to step-by-step reasoning verification loops. This demonstrates that overcoming semantic inertia requires significantly higher inference-time compute to inhibit visual priors.}
    \label{fig:case}
\end{figure*}

See \cref{fig:case}.

\subsection{\texorpdfstring{\ac{lcv}}{} Solution Sample}

\begin{Verbatim}[breaklines=True]
YOU_CHARS    = {}
STOP_CHARS   = {}
WIN_CHARS    = {}
DEFEAT_CHARS = {}
PUSH_CHARS   = {}
SINK_CHARS   = {}
MELT_CHARS   = {}
HOT_CHARS    = {}
DANGEROUS_TEXT_CHARS = {sorted(list(dangerous_text_chars))}

def next_state(grid, move):
    height = len(grid)
    if height == 0: return grid
    width = len(grid[0])

    new_grid = [row[:] for row in grid]

    directions = {{
        "UP": (-1, 0), "DOWN": (1, 0), "LEFT": (0, -1), "RIGHT": (0, 1)
    }}
    if move not in directions: return new_grid
    dy, dx = directions[move]

    # 1. Find YOU
    you_pos = []
    for r in range(height):
        for c in range(width):
            cell = new_grid[r][c]
            # Fast check
            has_you = False
            for char in cell:
                if char in YOU_CHARS:
                    has_you = True
                    break
            if has_you:
                for char in cell:
                    if char in YOU_CHARS:
                        you_pos.append((r, c, char))
    
    if not you_pos: return new_grid

    # 2. Move YOU
    # Sort to handle multiple YOU consistently
    you_pos.sort() 
    
    for r, c, me in you_pos:
        if me not in new_grid[r][c]: continue

        nr, nc = r + dy, c + dx

        # Map Boundary for YOU
        if not (0 <= nr < height and 0 <= nc < width):
            continue

        target_cell = new_grid[nr][nc]
        
        # --- PUSH Logic ---
        has_push = False
        for obj in target_cell:
            if obj in PUSH_CHARS:
                has_push = True
                break
        
        if has_push:
            chain = []
            curr_r, curr_c = nr, nc
            can_push = True
            
            while True:
                # [Critical] Boundary Check for PUSH Chain
                if not (0 <= curr_r < height and 0 <= curr_c < width):
                    can_push = False; break
                
                cell_objs = new_grid[curr_r][curr_c]
                push_objs_here = [o for o in cell_objs if o in PUSH_CHARS]
                
                if not push_objs_here:
                    # End of chain. Check BLOCKING.
                    is_blocked = False
                    for o in cell_objs:
                        if o in STOP_CHARS:
                            is_blocked = True; break
                    if is_blocked: can_push = False
                    break 
                else:
                    chain.append((curr_r, curr_c))
                    curr_r += dy
                    curr_c += dx
            
            if not can_push:
                continue 

            # Execute Push (Reverse Order)
            for tr, tc in reversed(chain):
                n_tr, n_tc = tr + dy, tc + dx
                
                src_cell = new_grid[tr][tc]
                moving = [o for o in src_cell if o in PUSH_CHARS]
                staying = "".join([o for o in src_cell if o not in PUSH_CHARS])
                
                new_grid[tr][tc] = staying
                new_grid[n_tr][n_tc] += "".join(moving)

            # Refresh target_cell after push
            target_cell = new_grid[nr][nc]

        # --- STOP Logic (Final check) ---
        is_blocked = False
        for obj in target_cell:
            if obj in STOP_CHARS:
                is_blocked = True; break
        if is_blocked: continue

        # --- Dangerous Text Logic (NEW!) ---
        is_text_dead = False
        for obj in target_cell:
            if obj in DANGEROUS_TEXT_CHARS:
                is_text_dead = True; break
        if is_text_dead:
            new_grid[r][c] = new_grid[r][c].replace(me, "", 1)
            continue

        # --- Enter Logic ---
        is_dead = False
        for obj in target_cell:
            if obj in DEFEAT_CHARS:
                is_dead = True; break
        if is_dead:
            new_grid[r][c] = new_grid[r][c].replace(me, "", 1)
            continue

        is_sink = False
        sink_obj = None
        for obj in target_cell:
            if obj in SINK_CHARS:
                is_sink = True; sink_obj = obj; break
        if is_sink:
            new_grid[r][c] = new_grid[r][c].replace(me, "", 1)
            new_grid[nr][nc] = new_grid[nr][nc]. \\
            replace(sink_obj, "", 1)
            continue
        
        is_melt_hot = False
        if me in MELT_CHARS:
            for obj in target_cell:
                if obj in HOT_CHARS:
                    is_melt_hot = True; break
        if is_melt_hot:
            new_grid[r][c] = new_grid[r][c].replace(me, "", 1)
            continue

        # Normal Move
        new_grid[r][c] = new_grid[r][c].replace(me, "", 1)
        new_grid[nr][nc] += me

    return new_grid

def check_win(grid):
    height = len(grid)
    width = len(grid[0])
    for r in range(height):
        for c in range(width):
            cell = grid[r][c]
            if not cell: continue
            has_you = False
            has_win = False
            for char in cell:
                if char in YOU_CHARS: has_you = True
                if char in WIN_CHARS: has_win = True
            if has_you and has_win:
                return True
    return False
"""
\end{Verbatim}

\section{Further Experiments}

\subsection{Ablation I: Training and Inference Efficiency}\label{sec:supp:efficiency}

We give an emparical comparison about the computational cost of our approach against both iterative reasoning baselines and state-of-the-art proprietary models in \cref{tab:efficiency}. All local training and inference experiments, including our \ac{lcv} model and the Qwen/TheoryCoder baselines, were conducted on a single node equipped with 8 $\times$ NVIDIA RTX 3090 (24GB) GPUs. Proprietary models (Gemini and GPT-4o) were evaluated via official APIs.

The results demonstrate the relative efficiency of the \ac{lcv} strategy. To solve the required code generation tasks, Gemini 3 Pro requires over 3k tokens of reasoning to navigate the logical conflict. Similarly, TheoryCoder scales linearly with problem complexity. In contrast, the \ac{lcv} agent requires only a minimal training investment ($\approx$ 600 paired samples for 5 epochs, costing about 30 minutes in one computing node). Once trained, it generates the correct world model in a single, compact pass ($\approx$ 800 tokens) on local hardware, reducing inference latency by approximately $4\times$ to $7\times$ compared to CoT-based approaches. This confirms that the contrastive objective successfully transforms the complex cognitive task of inhibition into an efficient, retrieved reflex.

\begin{table}[t!]
    \centering
    \small
    \caption{\textbf{Efficiency Comparison of Code-Based Methods.} While large foundation models (Qwen) and reasoning models (Gemini) rely on heavy test-time compute (either via parameter count or token volume), \ac{lcv} utilizes a small \ac{sft} set ($\sim$600 samples) to enable rapid, single-pass world modeling on local hardware.}
    \label{tab:efficiency}
    \resizebox{\linewidth}{!}{%
        \begin{tabular}{lcccc}
            \toprule
            \textbf{Method} & \textbf{Inference Mode} & \textbf{Avg Tok.} & \textbf{Time (s)} \\
            \midrule
            Qwen2.5-72B-Instruct & Direct Ans & $\sim$1k & 42.5 \\
            Gemini 3 Pro Preview & Think + Ans & $\sim$3k & 35.0 \\
            TheoryCoder & Iterative Loop & $\sim$3k & 65.2 \\
            \midrule
            \textbf{LCV (Ours)} & \textbf{\acs{sft} + Ans} & \textbf{$\sim$1k} & \textbf{8.4} \\
            \bottomrule
        \end{tabular}%
    }%
\end{table}

\subsection{OOD Generalization Robustness}\label{sec:supp:generalization}

To test the generalization capability of different baselines, we define three specific evaluation protocols:

\begin{enumerate}
    \item In-Distribution (In-Dist.): The test set follows the same data distribution as the training set, serving as the standard benchmark.
    \item Map Generalization (Map Gen.): Models are evaluated on novel spatial environments and map layouts that were not seen during training. This tests the model's ability to decouple logical rules from specific spatial configurations.
    \item Combination Generalization (Combo Gen.): This setting introduces novel combinations of rules and logical constraints, creating a systematic distribution shift that tests compositional generalization.
\end{enumerate}

We employ a Map Generalization set consisting of 30 unseen spatial layouts to assess whether the model overfits to specific map configurations. Furthermore, to test compositional reasoning, we construct a Combination Generalization split that introduces three systematic rule changes unseen during training: \textbf{Water is Push}, \textbf{Rock is Defeat}, and \textbf{Wall is Melt}.

\begin{table}[t!]
    \centering
    \small
    \caption{\textbf{Generalization Robustness.} Success Rate (SR) across generalization splits. }
    \label{tab:generalization}
    \resizebox{\linewidth}{!}{%
        \begin{tabular}{lccc}
            \toprule
            \textbf{Method} & \textbf{In-Dist.} & \textbf{Map Gen.} & \textbf{Combo Gen.} \\
            \midrule
            Qwen2.5-72B-Instruct    & 52.14\% & 53.40\% & 48.22\% \\
            Gemini 3 Pro Preview    & 42.85\% & 42.60\% & 39.62\% \\
            TheoryCoder             & 55.71\% & 53.72\% & 50.68\% \\
            \midrule
            \textbf{LCV (Ours)}    & 76.42\% & 74.28\% & 72.36\% \\
            \bottomrule
        \end{tabular}%
    }%
\end{table}

As shown in \cref{tab:generalization}, \ac{lcv} significantly outperforms baselines across all metrics. A potential concern regarding \ac{lcv} is its generalization capability; unlike generalist \acp{llm} (\eg, Qwen, Gemini) that leverage vast pre-trained parametric knowledge, \ac{lcv} relies on fine-tuning, which can theoretically increase the risk of overfitting. However, our statistical analysis refutes this. In the Map Gen. setting, we observed no statistically significant difference (p>0.05) between the In-Distribution and Map Generalization scores, quantitatively verifying that \ac{lcv} is robust to spatial variations.

Regarding the Combo Gen. split, while all methods experience a performance drop, we attribute this to the systematic deviation introduced by the new rules. To validate this, we compared the magnitude of the performance drop of \ac{lcv} against that of the baselines. We also found no statistically significant difference in the degradation rates (p>0.05). This confirms that the drop is a natural consequence of the increased logical complexity shared across all models, rather than a weakness specific to our approach. Crucially, \ac{lcv} maintains a substantial lead ($\sim$20\%), proving its reasoning framework remains robust even when facing these novel logical constraints.

\begin{figure}[ht!]
    \centering
    \begin{tabular}{cc}
        \includegraphics[width=0.42\linewidth]{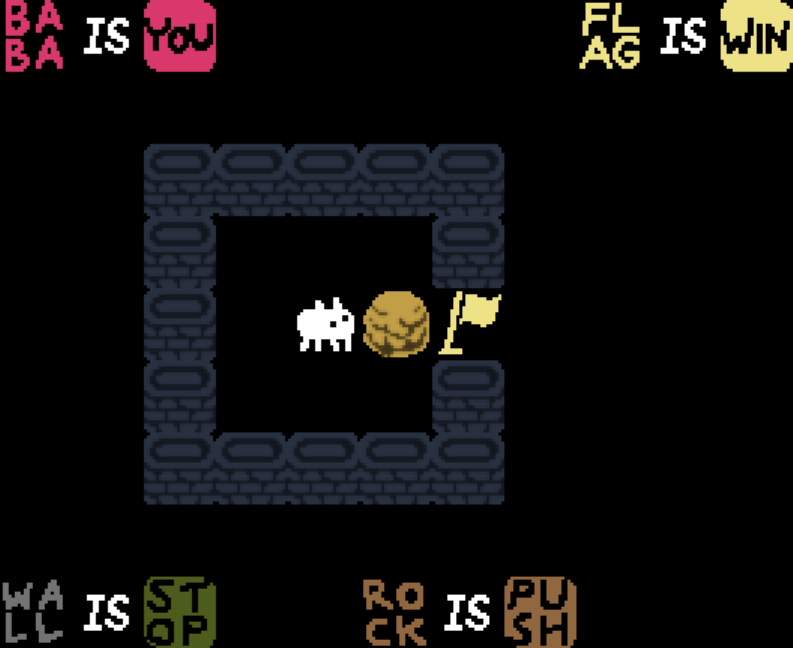} &
        \includegraphics[width=0.51\linewidth]{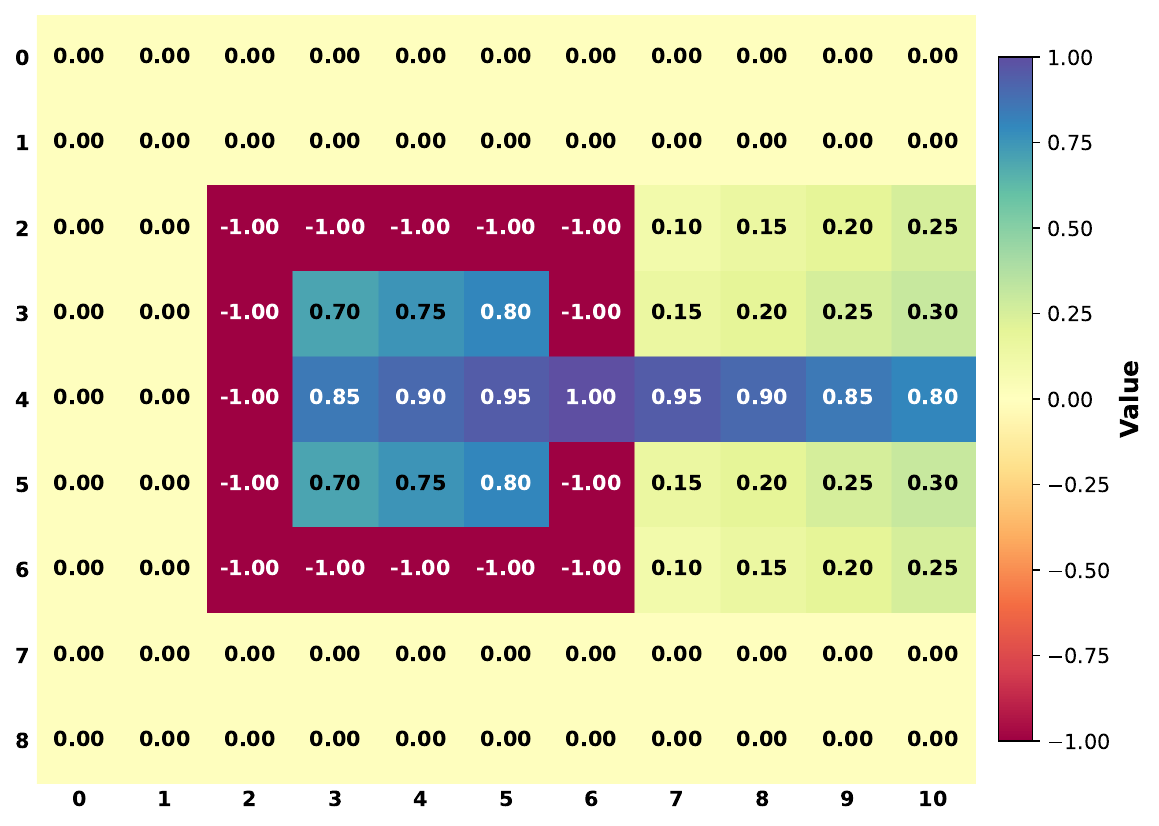} \\
        (\textbf{a}) State 1 & (\textbf{b}) Q-value 1 \\
        \includegraphics[width=0.42\linewidth]{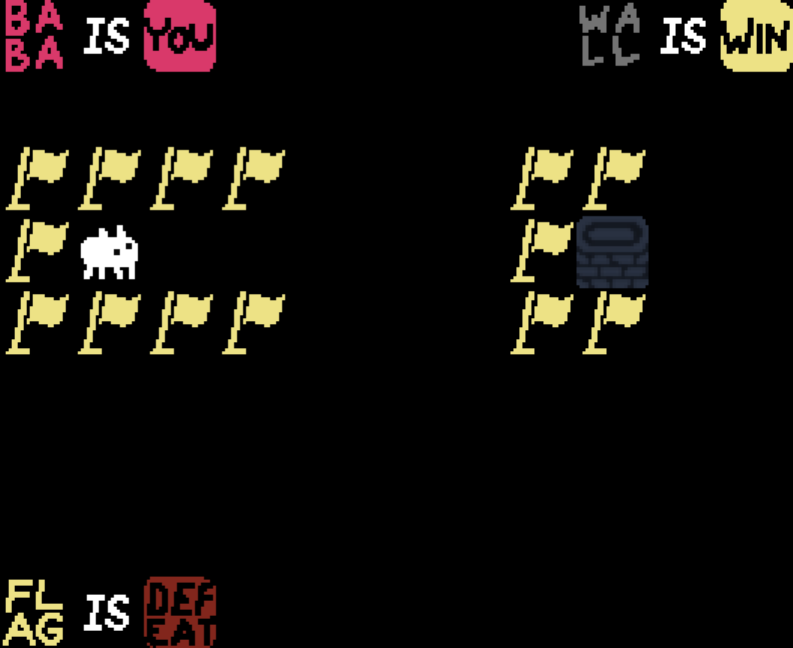} &
        \includegraphics[width=0.51\linewidth]{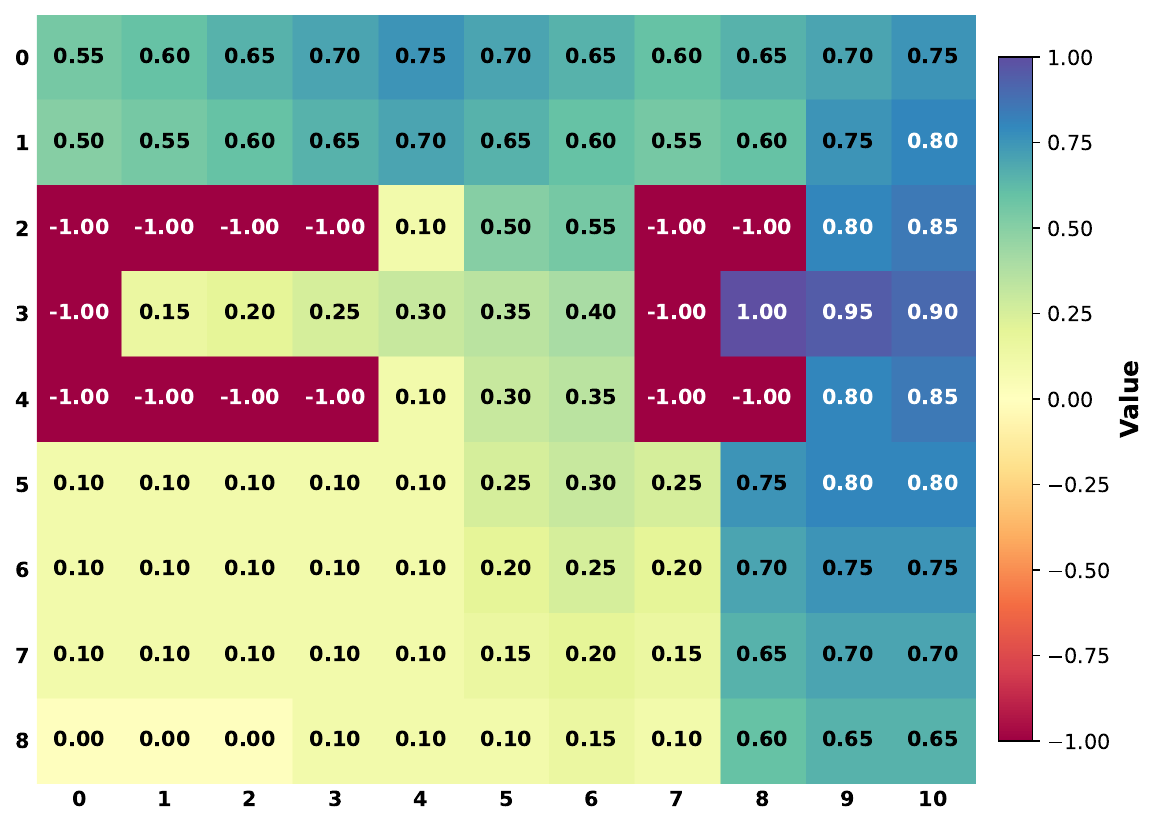} \\
        (\textbf{c}) State 2 & (\textbf{d}) Q-value 2 \\
        \includegraphics[width=0.45\linewidth]{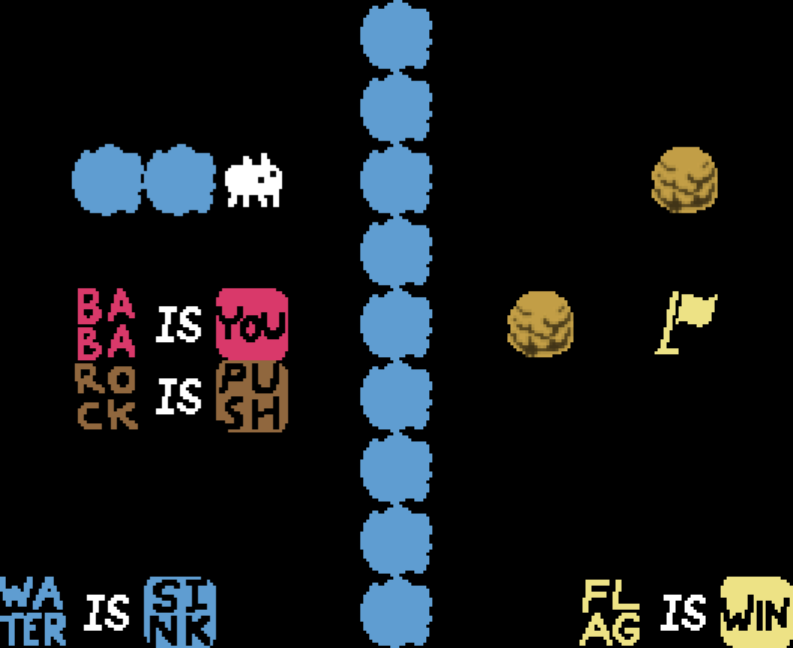} &
        \includegraphics[width=0.55\linewidth]{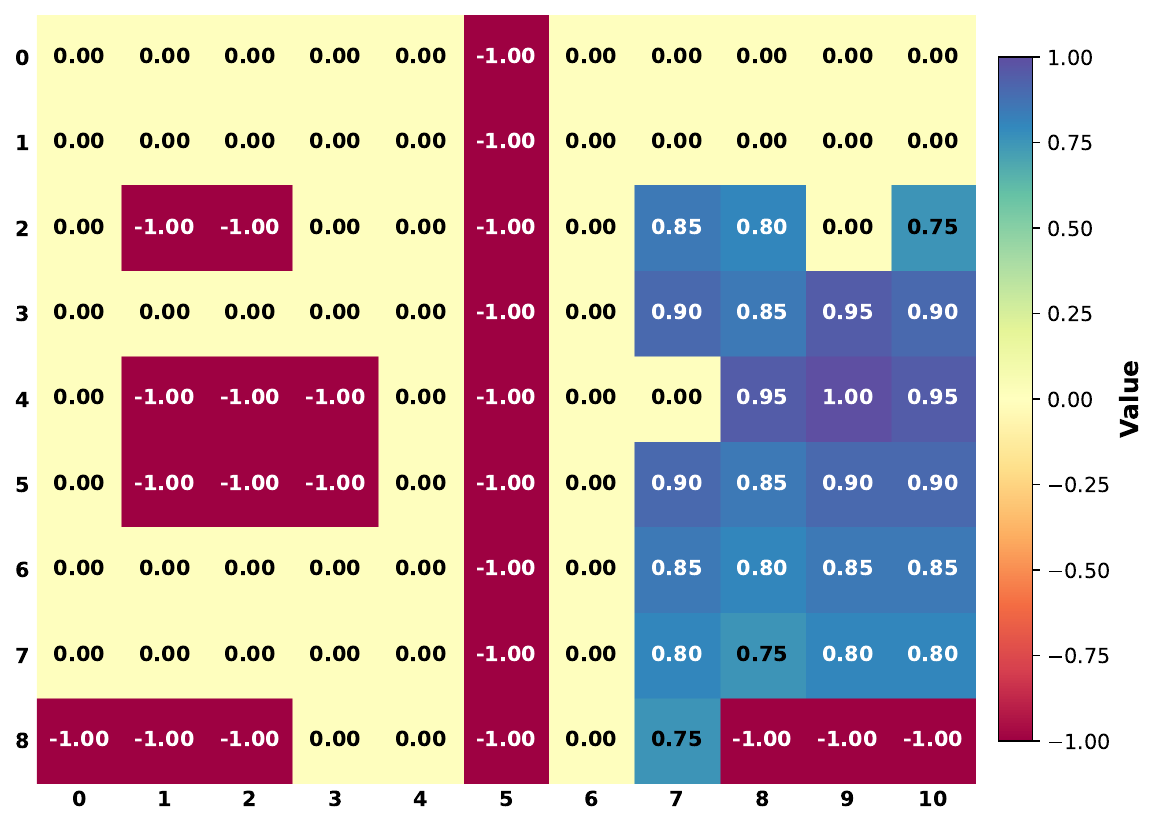} \\
        (\textbf{e}) State 3 & (\textbf{f}) Q-value 3 \\
    \end{tabular}
    \caption{\textbf{Q-value Visualizations.} For each scenario, the left panel shows the environment state, and the right panel shows the model's predicted $Q$-values for all possible actions in that state. From top to bottom: semantic alignment (Tier 1), semantic conflict (Tier 2), and dynamic rule change (Tier 3).}
    \label{fig:qvalue_all}
\end{figure}

\subsection{Q-Value Visualization for Model Reasoning}

To further understand the model's reasoning and decision-making process, we also recorded the model's predicted $Q$-values (action-value functions) at each inference step. As shown in \cref{fig:qvalue_all}, the left column contains visualizations of representative game states, while the right column shows the corresponding $Q$-value distributions over possible actions in those states. We present examples covering aligned semantics, semantic conflict, and dynamic rule changes.

As shown in \cref{fig:qvalue_all}, for Tier 1 (semantic alignment), the model's $Q$-value distribution is intuitive, preferring optimal actions that match commonsense reasoning (e.g., moving directly toward the goal while avoiding walls). In Tier 2 (semantic conflict, e.g., ``WALL IS YOU''), baselines without explicit code grounding still assign low or even negative $Q$-values to actions that violate their learned priors, reflecting an inability to inhibit semantic inertia. In contrast, our \ac{lcv} model, trained with contrastive objectives, correctly identifies that interacting with the wall is in fact desirable under the current rules and assigns higher $Q$-values to these actions. In Tier 3 (dynamic rule change), generic LLM-based agents often display significant $Q$-value instability and lag in adapting to new physics. 

\section{Declaration}

AI tools were used solely for grammar checking and language polishing for the overall paper; all ideas, technical content, and substantive writing are entirely the authors' own.

All code and benchmark data used in this work are released under the MIT License, allowing free use, modification, and distribution for research and non-commercial purposes.

We confirm that all existing artifacts used in this work were employed in accordance with their intended use and licensing terms. For the artifacts we created, we specify that they are intended strictly for research purposes, and their release is fully compatible with the original access conditions of any resources on which they are based.
\end{document}

%% file: config.tex
\usepackage[T1]{fontenc}
\usepackage[utf8]{inputenc}
\usepackage{times,latexsym,microtype,inconsolata}
\usepackage{graphicx} \graphicspath{{figures/}}
\usepackage{amsmath,amssymb,mathabx,mathtools,amsthm,nicefrac}
\usepackage[capitalise,nameinlink]{cleveref}
\usepackage[skip=3pt,font=small]{subcaption}
\usepackage[skip=3pt,font=small]{caption}
\usepackage[printonlyused]{acronym}
\usepackage[linesnumbered,ruled,vlined]{algorithm2e}
\usepackage{wrapfig}
\usepackage[dvipsnames,svgnames,x11names,table]{xcolor}
\usepackage{enumerate}
\usepackage{xspace}
\usepackage[misc]{ifsym}
\usepackage{verbatim,fancyvrb,fvextra,listings}
\usepackage{booktabs,tabularx,colortbl,multirow,multicol,array,makecell,tabularray}
\usepackage{siunitx}

\makeatletter
\DeclareRobustCommand\onedot{\futurelet\@let@token\@onedot}
\def\@onedot{\ifx\@let@token.\else.\null\fi\xspace}
\def\eg{\emph{e.g}\onedot}

\def\vs{\emph{vs}\onedot}

\makeatother

\acrodef{llm}[LLM]{Large Language Model}
\acrodef{nlp}[NLP]{Natural Language Processing}
\acrodef{cot}[CoT]{Chain-of-Thought}
\acrodef{lcv}[LCV]{Code-Grounded Vistas}
\acrodef{pal}[PAL]{Program-Aided Language models}
\acrodef{sft}[SFT]{Supervised Fine-Tuning}
\acrodef{pddl}[PDDL]{Planning Domain Definition Language}
\newcommand{\benchmark}{\textsc{BabaBench}\xspace}
\newcommand{\game}{\textit{Baba Is You}\xspace}

\SetKwInput{Input}{Input}
\SetKw{Continue}{continue}